\documentclass[letterpaper, 10 pt, conference]{ieeeconf}  

\IEEEoverridecommandlockouts                              

\usepackage{cite}
\usepackage{amsmath,amssymb,amsfonts}
\usepackage{wasysym}
\usepackage{graphicx}
\usepackage{textcomp}
\usepackage[dvipsnames]{xcolor}
\usepackage{multirow}
\usepackage{enumerate}
\usepackage{hyperref}
\usepackage{comment}
\usepackage{url} 
\usepackage[export]{adjustbox}
\usepackage[nolist,nohyperlinks]{acronym}
\usepackage{algorithm,algpseudocode}

\renewcommand{\baselinestretch}{0.933} 
\DeclareUnicodeCharacter{2212}{\textendash}

\title{\LARGE \bf
Object Augmentation Algorithm: Computing virtual object motion and object induced interaction wrench from optical markers
}

\author{Christopher Herneth$^{1}$, Junnan Li$^{1}$, Muhammad Hilman Fatoni$^{1}$, Amartya Ganguly$^{1}$, and Sami Haddadin$^{1}$
\thanks{$^{1}$Christopher Herneth, Junnan Li, Muhammad Hilman Fatoni, Amartya Ganguly, and Sami Haddadin are with Chair of Robotics and Systems Intelligence, MIRMI - Munich Institute of Robotics and Machine Intelligence, Technical University of Munich (TUM), Munich, Germany {\tt\footnotesize \{christopher.herneth, junnan.li, mhilman.fatoni, amartya.ganguly, haddadin\}@tum.de}} %
}

\begin{document}
\bibliographystyle{unsrt}

\maketitle
\thispagestyle{empty}
\pagestyle{empty}


\begin{abstract}

This study addresses the critical need for diverse and comprehensive data focused on human arm joint torques while performing activities of daily living (ADL). Previous studies have often overlooked the influence of objects on joint torques during ADL, resulting in limited datasets for analysis. To address this gap, we propose an Object Augmentation Algorithm (OAA) capable of augmenting existing marker-based databases with virtual object motions and object-induced joint torque estimations. 
The OAA consists of five phases: (1) computing hand coordinate systems from optical markers, (2) characterising object movements with virtual markers, (3) calculating object motions through inverse kinematics (IK), (4) determining the wrench necessary for prescribed object motion using inverse dynamics (ID), and (5) computing joint torques resulting from object manipulation. The algorithm's accuracy is validated through trajectory tracking and torque analysis on a 5+4 degree of freedom (DoF) robotic hand-arm system, manipulating three unique objects.
The results show that the OAA can accurately and precisely estimate 6 DoF object motion and object-induced joint torques. Correlations between computed and measured quantities were $>$ 0.99 for object trajectories and $>$ 0.93 for joint torques. The OAA was further shown to be robust to variations in the number and placement of input markers, which are expected between databases. Differences between repeated experiments were minor but significant ($p < 0.05$).
The algorithm expands the scope of available data and facilitates more comprehensive analyses of human-object interaction dynamics.

\end{abstract}



\section{Introduction} 
High-quality data, characterised by comprehensive datasets and reproducible methodology for grasping related tasks performed by humans alone and with assistive systems, is scarce. This scarcity primarily stems from the challenges associated with collecting such data from human subjects, often requiring expensive equipment that is not commonly available. Consequently, data reported in the literature is frequently collected for narrow and specialised use cases \cite{li2023gendexgrasp, wang2023dexgraspnet, Olugbade2022_HumanMovementDatabases}. As a result, researchers seeking a robust data foundation for their investigations face the challenge of searching for and combining multiple accessible datasets or relying on singular studies.
 However, there are notable exceptions where robust and reproducible hand kinematics and grasping datasets exist \cite{akhbari2019predicting, jarque2020large}.

A marked example of this scarcity is the lack of diverse data on human arm joint torques during \ac{adl}, with the authors being only aware of a single study \cite{Rosena_ADL_torque} covering a limited set of \ac{adl}. Over the past decades, their data has been consistently used for various fields such as rehabilitative robotics \cite{Toedtheide2023_ExoPro, Buongiorno2018}, data-driven design optimisation \cite{Fennel2022, Sha2020}, and continual biomechanical investigations of \ac{adl} manipulation \cite{Adams2015, Udawatta2019}. The manipulation of objects forms an integral task of \ac{adl}, where object manipulation-induced joint torques are at times multiple times higher than limb-related terms. However, the influence of objects was neglected in \ac{adl} joint torque estimations of \cite{Rosena_ADL_torque}. The lack of representative \ac{adl} joint kinetics led the authors of Damerla et al.  \cite{Damerla2022_2Rod_Parallel} and Toedtheide et al. \cite{AlexanderToedtheide2022_MIRMI_transhumeralProsthesis} to dimension their upper limb prosthetic mechanisms based on the peak human torque capabilities reported in \cite{Sasaki2010_WT(angle)}. However, the maximal torques generated by human joints exceed functional requirements, resulting in excessively heavy prosthetic devices — a significant concern for users \cite{Biddiss2007_25Years_abbandonment_studyreview, Davidson2002_Austr_Prost_Use, Cordella2016_Need_ULPU}.

Introducing objects into the dynamic analysis of human trials was discussed extensively based on five different methods in \cite{Akhavanfar2022_2HandSimCompare}. The authors concluded that approaches based on wrenches, which rely on known object kinematics, perform better when compared to procedures that involve adding object mass and inertia properties to the hand. The former approach resulted in overall reduced dynamic inconsistencies during inverse dynamics. In Muller et al. \cite{Muller2022_pointMass_torqueestimation}, object locations were inferred from inertial motion captured hand kinematics. However, the modelling of objects as a point mass did not accurately represent object dynamics, causing errors in their L5/S1 moment estimations, which increased with object mass and inertia. Hence, accurate object motions, leading to an object wrench $\in \mathbb{R}^6$, are required for reliable object-induced joint torque computations. Wrench-based methods enable the investigation of complex manipulation scenarios where objects are constrained by the environment (i.e., partial gravity compensation, reduced \ac{dof} of object motion, friction, etc.). Unfortunately, datasets jointly reporting human and object kinematics are few, with \cite{Krebs2021_ADL_KIT_Bimanual, Mandery2015_ADL_KIT_WholeBody} forming noteworthy exceptions. Recent databases such as \cite{Liang2020_ASIAN_ADL}, and \cite{SenayMihcin2023_ADL_PrevioslyExcludedADL}, specifically cover Asian demographic, and previously excluded \ac{adl} underlining that no singular database exists that can satisfy all purposes. 

Enabling the expansion and diversification of \ac{adl} joint torque data, we propose an algorithm capable of automatically and reliably augmenting existing marker-based databases with virtual object motions and object-induced joint torque estimations. Contrary to previous studies \cite{Muller2022_pointMass_torqueestimation, OuadoudiBelabzioui2023_upperLimbJointLoadANN, Wang2021L5/S1MomentEstimationDuringLifting}, we capture 6 \ac{dof} object motion derived from hand kinematics embedded in associated markers. Arbitrary object dynamics are accurately represented by the wrench necessary to move virtual object representations along prescribed trajectories. Inverse dynamics and simplified, single-point contact modelling compute induced joint torques. We validate our method in physical grasping and manipulation experiments against measured ground truth quantities, demonstrating the algorithm's accuracy, precision, and robustness to variations between datasets. 

\label{sec: Introduction}
\begin{figure}[tb]
    \centering
    \fontsize{6.5pt}{6.5pt}\selectfont
    \def\svgwidth{0.49\textwidth}
    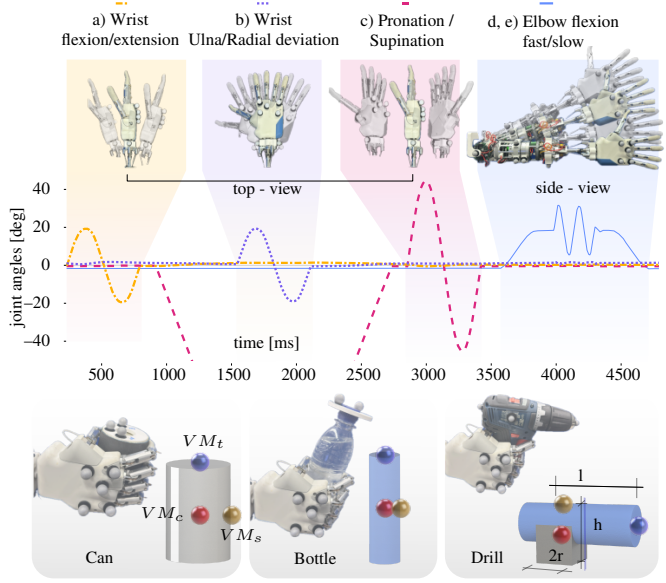
    \caption{Top Panel: 4 DoF limb motion trajectories and motion pictograms. Bottom panel: Grasped experimental objects, representative virtual objects, with and marker placements.}
    \label{fig: joint_traj}
\end{figure}

\begin{figure*}[tb]
    \centering
    \fontsize{6.5pt}{6.5pt}\selectfont
    \def\svgwidth{1\textwidth}
    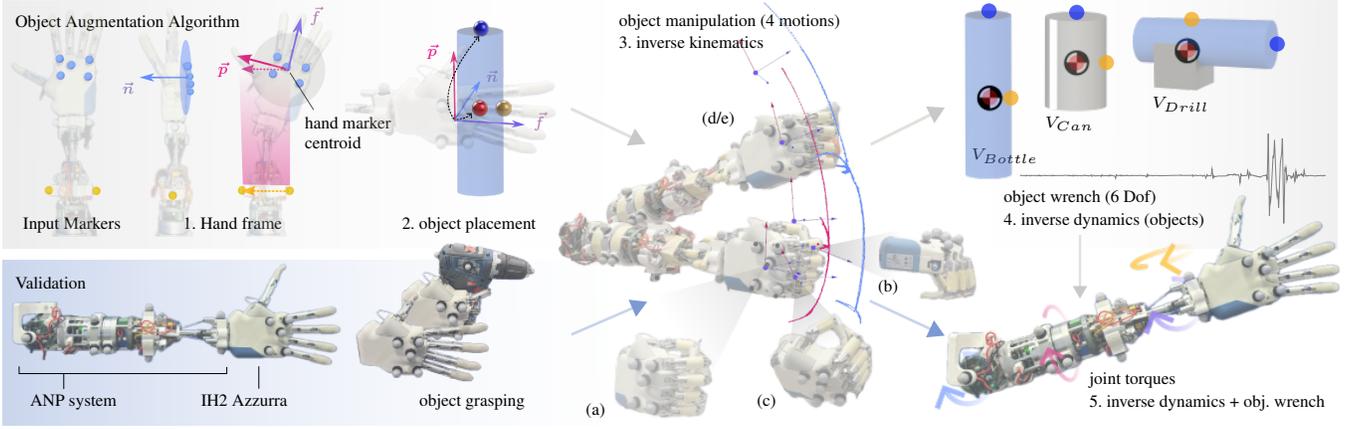
    \vspace{-0.6\baselineskip}
    \caption{The 5-phase computation pipeline of the Object Augmentation Algorithm: 1. hand frame from input markers, 2. virtual object placement in hand frame, 3. virtual object trajectory, 4. object wrench from object motion and 5. joint torques from object wrench. Validation: Object grasping - object manipulation and tracking - torque measurements.}
    \label{fig: algo}
    \vspace{-1.8\baselineskip}
\end{figure*}
\section{Methods and Materials} 
\label{sec: Methods}
\subsection{Measurement of ground truth data}
\subsubsection{Grasping experiments}
\label{grasp_exp}
A 4 \ac{dof} transhumeral prosthesis platform, the \ac{ANP}\cite{AlexanderToedtheide2022_MIRMI_transhumeralProsthesis} equipped with a 5 \ac{dof} robotic hand (IH2 Azzurra, Prensilia SRL, Italy) was used to perform grasping experiments. The kinematics of the \ac{ANP} and IH2 reflect those of the human elbow, forearm, wrist and hand. The objects utilized were a 425 ml tin can, a 500 ml water bottle, and a cordless drill weighing 720 g, Bosch GSR 12V-35 without battery. They were manually placed in the open robotic hand and grasped with the pre-programmed cylindrical grasp. Fig \ref{fig: joint_traj}. depicts corresponding joint configuration trajectories, where the ANP was mounted on a static fixture at the elbow joint. Subsequently, objects were moved by the ANP along four pre-programmed motions as follows:

\begin{itemize}
\item Motion 1, Fig. \ref{fig: joint_traj} (a) shows sinusoidal wrist flexion/extension $\pm$ 20$^{\circ}$ with elbow flexed at 90$^{\circ}$
\item Motion 2, Fig. \ref{fig: joint_traj} (b) shows sinusoidal wrist deviation $\pm$ 20$^{\circ}$ while elbow is flexed at 90$^{\circ}$ with forearm pronated at 90$^{\circ}$.
\item Motion 3, Fig. \ref{fig: joint_traj} (c) shows sinusoidal pronation/supination of the forearm $\pm$ 45$^{\circ}$
\item Motion 4, Fig. \ref{fig: joint_traj} (d): elbow flexion/extension: $\pm$ 18$^{\circ}$ and Fig. \ref{fig: joint_traj} (e): fast sinusoidal elbow flexion/extension: 18$^{\circ}$ $\pm$ 14$^{\circ}$
\end{itemize}

After each motion, objects were removed from the grasp. Experiments were repeated three times per object. Trajectories were tracked by an impedance controller set to a stiffness of 10 Nm/rad and a damping of 0.7 Nms. Ground truth recordings were obtained for joint angle, velocity, and torque from sensors of the \ac{ANP}. 

\subsubsection{Recording of object trajectories using motion capture}

The ANP was affixed with seven reflective markers, each 12 mm in diameter. Five such markers were placed on the dorsal side of the robotic hand, which form the set $M_h$= $\{$$M_{h1}$, $M_{h2}$, $M_{h3}$, $M_{h4}$, $M_{h5}$$\}$ and 2 were placed on the wrist Ulnar and Radial styloids forming $M_w$= $\{$$US$, $RS$$\}$. Finally, each object also had three markers affixed such that both the ANP as well as the objects can be tracked during experiments. The motion capture was performed with seven Valkyrie cameras (Vicon Motion Systems Ltd, Oxford, UK) which was synchronised with the ANP using a user datagram protocol (UDP) system. 

\subsubsection{Dynamics of the system}

Considering the arm-hand system of the \ac{ANP} with four joints and their configuration $\boldsymbol{q}$, the dynamics of the system was formulated as 

\begin{equation}
    \boldsymbol{\tau} = \boldsymbol{\tau}_{P} + \boldsymbol{\tau}_{obj} = \boldsymbol{M}( \boldsymbol{q}) \boldsymbol{\ddot{q}} + \boldsymbol{C(q, \dot{q})} + \boldsymbol{G(q)} + \boldsymbol{J(q)}^T \boldsymbol{W}_{obj},
    \label{equ: equations of motion}
\end{equation}

where, $\boldsymbol{\tau}_{P}$ and $ \boldsymbol{\tau}_{obj}$ represent prosthesis and object related torques and $\boldsymbol{M}$ denoting the inertia matrix, and $\boldsymbol{C}$, centrifugal and Coriolis, and $\boldsymbol{G}$ gravity torques. Joint torques caused by the wrench from manipulating an object along its prescribed trajectory were considered as the end-effector wrench vector, $\boldsymbol{W}_{obj} \in \mathbb{R}^6$. Therefore, joint torques originating from objects were calculated by multiplying the transpose of the Jacobian matrix $\boldsymbol{J(q)}^T$ with the object wrench $\boldsymbol{W}_{obj}$. Ground truth object-induced joint torques $ \boldsymbol{\tau}_{Gobj}$ were computed for each experiment as the difference between ANP torque measurements $ \boldsymbol{\tau}_{meas}$ and simulated, prosthesis-related joint torques $ \boldsymbol{\tau}_{P_s}$. The latter were calculated from prosthesis terms in \eqref{equ: equations of motion}, and filtered measurements of $\boldsymbol{q}, \boldsymbol{\dot q}, \boldsymbol{\ddot q}$. This was necessary since measured torques cannot distinguish between object and prosthesis-related terms.

\subsection{Object Augmentation Algorithm}
We now have ground truth data consisting of joint angles, velocity, and torque along with the marker trajectories from the robotic hand and objects. In this section we propose and validate an Object Augmentation Algorithm (OAA) that computes the prescribed movement of objects based on optical markers affixed on the robotic hand and wrist. Joint torques induced by object manipulation were calculated by applying the wrench necessary for object motion with respect to the end effector. The algorithm pipeline comprises of five phases (Fig. \ref{fig: algo}):

\begin{enumerate}
    \item robotic hand coordinate frames $F_{hand}$ from optical wrist and hand markers
    \item virtual object marker trajectories
    \item virtual object kinematics by \ac{IK}
    \item virtual object wrench by \ac{ID}
    \item virtual object induced joint torques from object wrench 
\end{enumerate} 


The algorithm requires the following inputs as outlined in pseudocode Algorithm \ref{alg: OAA}:

\begin{itemize}
    \item Optical markers: trajectories of markers $\in$ $M_w$, and $M_h$.
    \item Object: dimensions, mass, and inertia $\in \{D, m, I\}$, and a set of virtual object markers and their location within the hand.
    \item Limb: wrist deviation angles $\alpha$ for each input marker frame.
    \item Limb: a kinematic model of the limb manipulating the object, specifically its Jacobian $J$.
\end{itemize}

\setlength{\textfloatsep}{2pt}
\begin{algorithm}[tb] 
\caption{Object Augmentation} \label{alg: OAA}
\begin{algorithmic}[]
\Require{$\{M_w, M_h\}$ optical wrist and hand marker frames}
\Require{$\{D, m, I\}$} object dimensions, mass, and inertia
\Require{$c(D)$ virtual object marker coordinates in hand frames}
\Require{$\alpha$} wrist deviation angle trajectory
\Require{$J$ Jacobian of the object manipulator}
\Ensure{$P_{obj}$ object motion, $W_{obj}$ object wrench}
\Ensure{$\tau_{obj}$ object induced joint torques}
\State {$F_{hand} \gets []$ hand coordinate frames}
\For{each $M_{w,i}, M_{h,i}$} 
    \State {$F_{hand}$.insert($[\vec{n_i}, \vec{p_i}, \vec{f_i}] = f(\alpha_i, M_{w,i}, M_{h,i})$)}
\EndFor
\State {$M_o \gets []$ $n$ virtual object marker trajectories}
\For{each $F_{hand,i}$ in $F_{hand}$} 
    \State {$M_o$.insert($\{F_i c_n(D) \ \forall$ virtual markers $n \}$}
\EndFor
\State {$P_{obj} = IK(M_o)$} \Comment{Object inverse kinematics}
\State {$W_{obj} = ID(P_{obj}, m, I)$} \Comment{Object inverse dynamics}
\State {$\tau_{obj} = J(q)^T{W}_{obj}$} \Comment{Limb inverse dynamics}
\end{algorithmic}
\end{algorithm}

\begin{figure}[hb!]
    \centering
    \fontsize{6.5pt}{6.5pt}\selectfont
    \def\svgwidth{0.48\textwidth}
    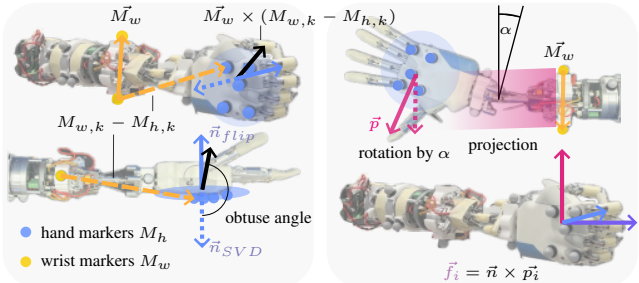
    \caption{Hand coordinate frame from hand $M_h$ and wrist $M_w$ markers.}
    \label{fig: handframe}
\end{figure}

Phase (1) computed the robotic hand coordinate systems $F_{hand,i} \in \{\vec{n_i}, \vec{p_i}, \vec{f_i}\}$ for each frame $i$ of input markers {$\{M_w, M_h\}$}, as shown in Fig. \ref{fig: handframe}. The first vector $\vec{n}$ represents the normal vector of the palm. It was calculated using the Singular Value Decomposition (SVD) method applied to $M_h$, where $\vec{n}$ was found as the right singular vector associated with the smallest singular value, equation \eqref{equ: n}. The intended direction of $\vec{n}$ depicted in Fig. \ref{fig: algo} is not guaranteed by SVD. The direction of $\vec{n_i}$ reverses if there is an obtuse angle between $\vec{n_i}$ and the result of the cross product between the vector $\vec{M_{w,i}}$ (unit vector from the Ulnar to the Radial styloid) and $M_{h,i,k}-M_{w,i,k}$ (the vector from any wrist to any hand marker, $k$ = arbitrary index) as shown in equation (\ref{equ: n})

\begin{equation}
\begin{aligned}
& U \Sigma [, , \vec{n_i}]^{T} = SVD(M_{h,i} - \overline{M_{h,i}}) \\
& s.t \ \ \vec{n_i} \cdot (\vec{M_{w,i}} \times (M_{h,i,k}-M_{w,i,k})) \leq 0.
\label{equ: n}
\end{aligned}
\end{equation}

$\vec{p}$, represents the thumbs up direction. It was realized by projecting $\vec{M_{w,i}}$ onto the plane defined by $\vec{n_i}$, and subsequent rotation around $\vec{n_i}$ by the deviation angle of the wrist in the current frame $\alpha_i$, equation \eqref{equ: p}:

\begin{equation}
    \vec{p_i} = R(\vec{n_i}, \alpha_i) (\vec{M_{w,i}} - (\vec{M_{w,i}} \cdot n_i)n_i).
    \label{equ: p}
\end{equation}

Finally, the robotic hand vector $\vec{f}$, representing the direction of the outstretched index finger, was formed by $\vec{n_i} \times \vec{p_i}$. The resulting coordinate system $F_{hand,i}$ was located at the robotic hand marker frame centroid ($\overline{M_{h,i}}$), which is the mean position of all markers on the dorsal side of the same robotic hand.


\begingroup

\setlength{\tabcolsep}{2pt} 

\begin{table}[ht]
\caption{Virtual object characteristics and associated virtual marker coordinates in $F_{hand}$}
\vspace{-1\baselineskip}
\label{tab: ObjChar}
\begin{tabular}{l|llllllllc}
\multirow{2}{*}{$V_{Object}$} & \multicolumn{1}{c}{$r$}  & \multicolumn{1}{c}{$h$ / $l$}  & \multicolumn{1}{c}{$m$} & \multicolumn{1}{c}{Ixx} & \multicolumn{1}{c}{Iyy} & \multicolumn{1}{c}{Izz} & \multicolumn{1}{c}{\multirow{2}{*}{VM}} &  \multicolumn{1}{c}{\multirow{2}{*}{$c$}}  &  \multicolumn{1}{c}{$\theta$}   \\
  & \multicolumn{2}{c}{{[}mm{]}}  & \multicolumn{1}{c}{{[}kg{]}} & \multicolumn{3}{c}{{[}kg m\textasciicircum{}2{]}}  &  &  &  \multicolumn{1}{c}{{[}deg{]}} \\ \hline
\multirow{2}{*}{$V_{Can}$}    & \multirow{2}{*}{37}          & \multirow{2}{*}{109}                                               & \multirow{2}{*}{0.51}        & \multirow{2}{*}{7e-4}   & \multirow{2}{*}{7e-4}   & \multirow{2}{*}{4e-4}   & $VM_t$                                          & $\vert r+H, h/2, r\vert^\top$  & 12                            \\
                        &                              &                                                                    &                              &                         &                         &                         & $VM_c$                                          & $\vert r+H, 0, r\vert^\top$  & 12                            \\ \hline
\multirow{2}{*}{$V_{Bottle}$} & \multirow{2}{*}{30}          & \multirow{2}{*}{210}                                                & \multirow{2}{*}{0.57}        & \multirow{2}{*}{2e-3}   & \multirow{2}{*}{2e-3}   & \multirow{2}{*}{3e-4}   & $VM_t$                                          & $\vert r+H, h/2, r\vert^\top$  & 12                            \\
                        &                              &                                                                    &                              &                         &                         &                         & $VM_c$                                          & $\vert r+H, 0, r\vert^\top$  & 12                            \\ \hline
\multirow{2}{*}{$V_{Drill}$} & \multirow{2}{*}{58}          & \multirow{2}{*}{\begin{tabular}[c]{@{}l@{}}88/\\ 128\end{tabular}} & \multirow{2}{*}{0.72}        & \multirow{2}{*}{3e-4}   & \multirow{2}{*}{1e-3}   & \multirow{2}{*}{1e-3}   & $VM_t$                                          & $\vert r+H, h, r+l\vert^\top$  & 0                             \\
                        &                              &                                                                    &                              &                         &                         &                         & $VM_c$                                          & $\vert r+H, h, r\vert^\top$  & 0                            
\end{tabular}
\end{table}
\endgroup

In phase (2), object movements were characterized by computing trajectories of virtual markers placed on the virtual objects $V_{obj} \in \{V_{can}, V_{bottle}, V_{drill}\}$. $V_{can}$ and $V_{bottle}$ were modelled as a cylinder with a radius $r$, height $h$, grasped by the robotic hand with a palm thickness of $H$, while the drill was modelled as cylinder resting on a cube, see Fig. \ref{fig: joint_traj}. Three virtual markers (VM) were placed on each of the $V_{obj}$, one on top centre of the cylinder ($VM_t$), the second on the object's \ac{CoM} ($VM_c$), and the third on the cylinder's circumference at the height of the \ac{CoM} ($VM_s$), as shown in Fig. \ref{fig: joint_traj}. The placement of the aforementioned virtual markers was according to $F_{hand}$ frame as described by \eqref{equ: A}, with VM coordinates of $VM_t$ and $VM_c$ collected in table \ref{tab: ObjChar}.

\begin{equation}
    M_{o} = \left\{
    \begin{vmatrix}
        1 & 0 & 0 \\
        0 & \cos(\theta) & -\sin(\theta)\\
        0 & \sin(\theta) & \cos(\theta) \\
    \end{vmatrix} 	
    \begin{vmatrix}
        \vec{n_i} & \vec{p_i} & \vec{f_i}
    \end{vmatrix} 	
    c \ \forall i
    \right\}\\
    \label{equ: A}
\end{equation}

Phase (3), calculated object motions $P_{obj}$ through \ac{IK} (OpenSim 4.4 \ac{IK} solver \cite{Delp2007_OpenSim}), where virtual marker trajectories $M_o$ were subjected to a 3rd order low pass filter (cutoff 6 Hz). Resulting motions were either $\in \mathbb{R}^6$ or span a subspace if environmental constraints exist (e.g., partial support by surfaces, joints, etc.). 

Phase (4), calculated the wrench $W_{obj} \in \mathbb{R}^6$ necessary to move objects along the prescribed motion by \ac{ID} (OpenSim 4.4 \ac{ID} tool) as well as dynamic object models. Object models were based on the real objects described in subsection \ref{grasp_exp}, with mass and inertia properties collected in table \ref{tab: ObjChar}.

Finally, in phase (5), joint torques resulting from object manipulation were computed by applying $W_{obj}$ to the hand. The interaction between the robotic hand and the object was assumed as a single point of contact at the center of mass (CoM) of the object. This resulted in 

\begin{equation}
    \tau_{obj} = {J(q)}^T {W}_{obj}.
    \label{equ: tau}
\end{equation}

Dynamic simulations of the robotic system were carried out in Matlab R2023b (The MathWorks Inc. 2023), with system dynamic properties estimated from CAD models. Friction torques were neglected in our analysis.

\subsection{Validation of OAA and statistical analysis}
\label{sec: OAA valid}
The validation of the OAA includes a comparison of measured object trajectories and joint torques from the \ac{ANP} with that of estimated by the OAA for the following scenarios:

\begin{enumerate}
    \item Computed vs. measured object trajectories 
    \item Computed vs measured object induced joint torques
    \item Sensitivity test: variations in the number and placement of hand markers $M_h$.
\end{enumerate}

The test for sensitivity was performed where the size of the hand marker set $\lvert M_h\rvert \in \{3, 4, 5, 6\}$ and placement of individual markers in $M_h$ on the dorsal side of the hand, were changed which altered the location of hand coordinate frames produced by the OAA. 

The average hand marker centroid distance for each $\lvert M_h\rvert$ was produced by changing the location of markers in the $M_h$ set a 1000 times, hence 4000 centroids were generated. The marker placement was randomised within a diameter of 6 cm, corresponding to the average breadth of the female hand \cite{AboulHagag2011_handsize} minus one marker diameter (1.2 cm) while ensuring a minimal inter-marker distance of 3 cm.

Differences between objects and repetitions were tested by the Kruskal-Wallis H-test, with post hoc analysis done with the Mann-Whitney U rank test. Values P $<$ 0.05 were considered statistically significant. Normality was tested by visual inspection of histograms. Pearson's $r$ was employed as a measure of correlation. The L1 (torques) and L2 (marker coordinates) norms and the \ac{RMS} error evaluated the distances between measured and computed quantities. Statistical analysis was performed using the Python package Scipy 1.10.0 (The SciPy community).

\section{Results} 
\label{sec: Results}
This section shows the performance and outcome of two cases, namely
\begin{enumerate}
    \item  virtual markers ($VM$) vs motion-captured markers ($GM$) 
    \item  virtual markers with offset ($VM_{off}$) vs motion capture markers ($GM$)
\end{enumerate}
\subsection{Sensitivity Analysis}

\begin{figure}[tb]
    \centering
    \fontsize{6.5pt}{6.5pt}\selectfont
    \def\svgwidth{0.59\textwidth}
    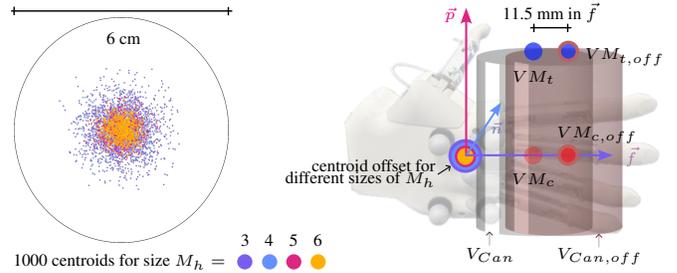
    \caption{Sensitivity analysis outcome of OAA, where centroids of marker clusters {3, 4, 5 and 6} were calculated within a 6 cm diameter. }
    \label{fig: sensitivity}
\end{figure}

Fig. \ref{fig: sensitivity} (left panel) shows 4000 centroids computed as in section \ref{sec: OAA valid}. Hand marker cluster sets sizes $\lvert M_h\rvert$ of 3, 4, 5, and 6, each contribute 1000 centroids. The mean distances between centroids were 14 mm for $\lvert M_h\rvert$=3, 12.4 mm for $\lvert M_h\rvert$=4, 10.4 mm for $\lvert M_h\rvert$=5, and 9.3 mm for $\lvert M_h\rvert$=6. The mean distance across all clusters sizes (11.5 mm) was used as the offset for case (2), where $VM_off$ = $VM$ + 11.5$\vec{f}$, as depicted for $V_{Can}$ in Fig. \ref{fig: sensitivity} (right panel). 

\subsection{Object trajectory tracking}

\begin{figure*}[tb]
    \centering
    \fontsize{6.5pt}{6.5pt}\selectfont
    \def\svgwidth{0.99\textwidth}
    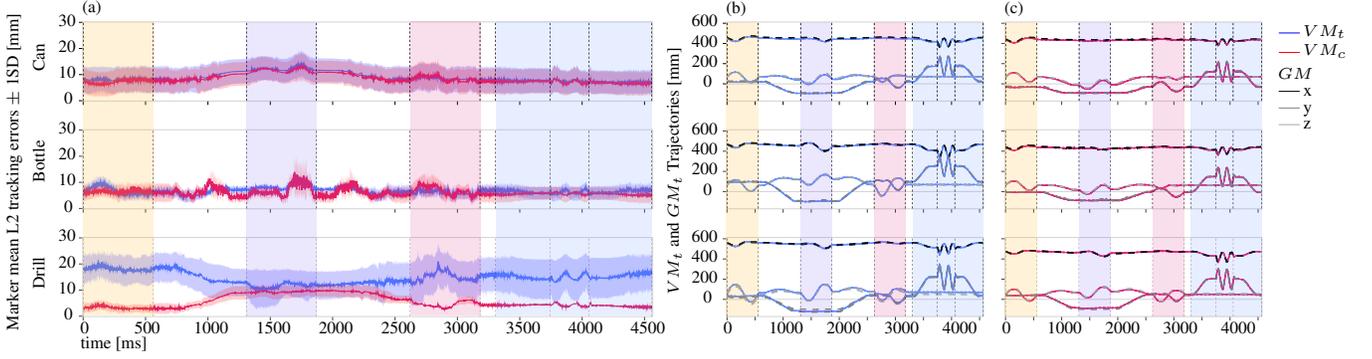
    \vspace{-0.6\baselineskip}
    \caption{(a) For case 1 -- 
 point-wise L2 tracking errors between virtual ($VM$) and ground truth markers ($GM$). Mean errors are solid lines. Error bands are $\pm$ 1SD across three repetitions. Panels (b) and (c) show x, y, and z marker trajectories affixed onto each object. Colored lines are $VM$ case 1 trajectories. $GM$ trajectories are grey-scale dashed. Vertical, coloured bands indicate joint motions.}
    \label{fig: marker error}
    \vspace{-1.8\baselineskip}
\end{figure*}

Table \ref{tab: Stat_ObjectTraj} collects error statistics markers ($VM$, $VM_{off}$ vs ground truth) averaged over three repetitions per object. Fig. \ref{fig: marker error} (a) shows marker trajectory errors as L2 distances between virtual and ground truth marker coordinates.  

\begingroup

\setlength{\tabcolsep}{2pt} 

\begin{table}[ht]
\scriptsize
\caption{Descriptive statistics of virtual vs. optical marker trajectory tracking errors over three repetitions per trial.}
\vspace{-1\baselineskip}
\label{tab: Stat_ObjectTraj}
\begin{tabular}{ll|llllll}
Object &  Marker & Mean$\pm$SD  &   $\overline{Max}$ &     $\overline{RMS}$ & $\overline{Corr_x}$ & $\overline{Corr_y}$ & $\overline{Corr_z}$  \\
 &   & [mm]  &  Mean$\pm$SD & Mean$\pm$SD &  &  &\\
 & & & [mm] & [mm] &  &  &  \\ \hline
 \multirow{4}{*}{Can} &     $VM_t$ &  9 $\pm$ 5 & 15 $\pm$ 5 &  9 $\pm$ 5 &   1.00 &   0.98 &   1.00 \\
 & $VM_{t, off}$ &  8 $\pm$ 4 & 13 $\pm$ 5 &  8 $\pm$ 4 &   1.00 &   0.99 &   1.00 \\
 &     $VM_c$ &  8 $\pm$ 5 & 14 $\pm$ 5 &  9 $\pm$ 4 &   0.99 &   0.99 &   1.00 \\
 & $VM_{c, off}$ &  7 $\pm$ 4 & 12 $\pm$ 5 &  7 $\pm$ 4 &   0.99 &   1.00 &   1.00 \\ \hline
 \multirow{4}{*}{Bottle} &     $VM_t$ &  7 $\pm$ 2 & 13 $\pm$ 1 &  7 $\pm$ 2 &   1.00 &   1.00 &   1.00 \\
 & $VM_{t, off}$ &  6 $\pm$ 2 & 13 $\pm$ 1 &  7 $\pm$ 1 &   1.00 &   1.00 &   1.00 \\
 &     $VM_c$ &  6 $\pm$ 2 & 17 $\pm$ 5 &  6 $\pm$ 2 &   0.99 &   0.99 &   1.00 \\
 & $VM_{c, off}$ &  6 $\pm$ 2 & 17 $\pm$ 5 &  6 $\pm$ 2 &   0.99 &   1.00 &   1.00 \\ \hline
\multirow{4}{*}{Drill} &     $VM_t$ & 15 $\pm$ 6 & 25 $\pm$ 6 & 15 $\pm$ 6 &   1.00 &   0.99 &   1.00 \\
 & $VM_{t, off}$ & 15 $\pm$ 6 & 25 $\pm$ 6 & 15 $\pm$ 6 &   1.00 &   0.99 &   1.00 \\
 &     $VM_c$ &  6 $\pm$ 1 & 10 $\pm$ 2 &  6 $\pm$ 1 &   1.00 &   1.00 &   1.00 \\
 & $VM_{c, off}$ &  6 $\pm$ 1 & 10 $\pm$ 2 &  6 $\pm$ 1 &   1.00 &   1.00 &   1.00 \\
\end{tabular}
\end{table}
\endgroup

$VM_t$ and $VM_c$ of the Can exhibited maximal L2 marker errors of $\sim$15$\pm$5 mm during wrist deviation. $VM_c$ tracking errors for the Bottle were maximal during wrist flexion, with broader error bands during forearm pronation and supination, 17$\pm$5 mm. In contrast, $VM_t$ error for the Bottle was smaller and narrowly distributed, 13$\pm$1 mm. Mean and \ac{RMS} errors were smaller than for the Can, 6$\pm$2 mm. In trials of the Drill, maximal $VM_t$ error was the largest between objects, 25$\pm$6. However, Drill $VM_c$ showed the smallest maximal error for all experiments, 10$\pm$2. Overall, wrist pronation increased errors for both Can $VM$, and the Bottle's $VM_c$. For the Drill, $VM_c$ errors increased while $VM_t$ errors were reported to be reduced. The error characteristics of $VM_s$ matched that of $VM_c$ for all objects.

Inter-repetition differences in cases (1) and (2) tracking errors were minor but significant $p < 0.05$ for the Can and the Bottle. However, they were negligible for the Drill.

Inter-case differences were found to be small but significant ($p < 0.05$) for each object. Trajectories for case (1) and case (2) (coloured) and ground truth  (grey-scale) x, y, and z coordinates are illustrated for all trials in Figs. \ref{fig: marker error} (b) for  $VM_t$, as well as Figs. \ref{fig: marker error}(c) for $VM_c$. Virtual and ground truth trajectories exhibited high correlation ($p << 0$) for all trials, resulting in nearly overlapping trajectories in Figs. \ref{fig: marker error} (b) and (c).

\subsection{Object induced joint torques}

The maximum marker errors reported by OpenSim for \ac{IK} were 92$\pm$36 $\mu$m for virtual and 118$\pm$19 $\mu$m  for ground truth markers, averaged over all trials. Joint torques reported below include prosthesis and object manipulation wrench-induced terms according to equation \eqref{equ: equations of motion}. Joints are labeled EF for the elbow flexion/extension, PS for forearm pronation/supination, WF for wrist flexion/extension, and WD for wrist Ulnar/Radial deviation. Torques were compared for cases:

\begin{enumerate}[I]
    \item $\boldsymbol{\tau}_{meas}$ = $\boldsymbol{\tau}_{P}$ + $\boldsymbol{\tau}_{obj}$ (torque measurements from the \ac{ANP} composed of torques from robot dynamics + torques induced by manipulating grasped objects)
    \item $\boldsymbol{\tau}_{P\_s}$ + $\boldsymbol{\tau}_{Vobj}$ (simulated prosthesis dynamics + object torques from virtual object trajectories)
    \item $\boldsymbol{\tau}_{P\_s}$ + $\boldsymbol{\tau}_{Vobj\_off}$ (simulated prosthesis dynamics + object torques from offset, virtual object trajectories)
    \item $\boldsymbol{\tau}_{P\_s}$ + $\boldsymbol{\tau}_{Gobj\_off}$ (simulated prosthesis dynamics + object wrench from ground truth, motion-captured object trajectories.)
\end{enumerate}

\begin{figure}[tb]
    \centering
    \fontsize{6.5pt}{6.5pt}\selectfont
    \def\svgwidth{0.47\textwidth}
    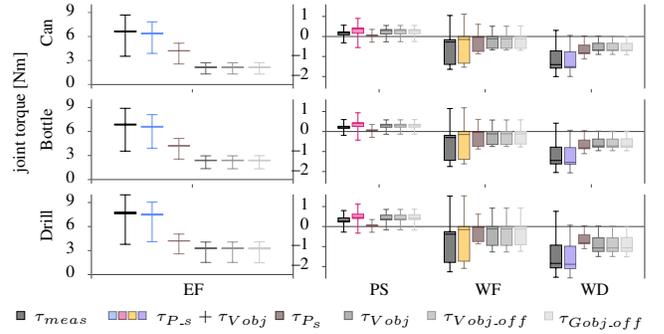
    \caption{Box-plots of different torque terms for each joint. Torques were computed over three repetitions per object.}
    \label{fig: torque boxplot}
\end{figure}

\begin{figure*}[tb]
    \centering
    \fontsize{6.5pt}{6.5pt}\selectfont
    \def\svgwidth{0.99\textwidth}
    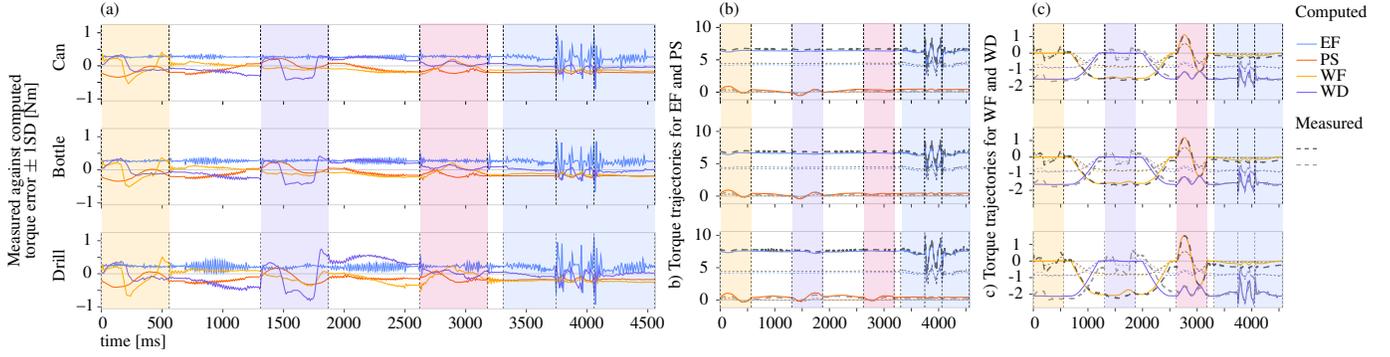
    \caption{Panel (a) -- Point-wise, L1 tracking errors between computed and measured joint torques. Mean errors are solid lines. Error bands are $\pm$ 1SD across three repetitions. Panel (b) and panel (c) -- Joint torque trajectories for each joint and the first trial of each object. Coloured, solid lines are computed joint torques, including virtual object wrenches. Coloured dotted lines are computed torques for prosthesis dynamics. Grey-scale dashed lines are measured joint torques, including objects. Grey-scale dotted lines are measured joint torques without grasped objects. Vertical, coloured bands indicate joint motions.}
    \label{fig: torque error}
\end{figure*}

Fig. \ref{fig: torque boxplot} reports joint torque distributions for cases I-IV. Black boxes indicate case I (measured) and coloured boxes case II (computed) joint torques, whereas brown boxes relate to simulated prosthesis torques in the absence of objects. Grey-scale boxes describe torque terms produced solely by the objects described in cases II-IV. Case I and II torques were similarly distributed for all joints except PS. In PS, the percentile and peak torques were higher in the computed case. In EF and WD joints, measured peak torques were higher than computed torques, while they were equal in the WF joint. Distributions of $\boldsymbol{\tau}_{Vobj}$, $\boldsymbol{\tau}_{Vobj\_off}$, and $\boldsymbol{\tau}_{Gobj\_off}$ were indistinguishable for all joints and objects. The ratio between $\boldsymbol{\tau}_{P\_s}$ and object-induced torques showed different qualities between joints. In EF, $\boldsymbol{\tau}_{P\_s}$ was higher than object terms, while they were lower in PS. For wrist joints, $\boldsymbol{\tau}_{P\_s}$ and object terms were distributed similarly. 

Fig. \ref{fig: torque error} (a) reports error trajectories$\pm$1SD between cases I and II, where errors were computed as point-wise L1 differences for each joint, averaged across repetitions of the same object. Since standard deviations are minor, no error bands are seen. Errors computed for offset and ground truth (case I vs case III) object trajectories nearly overlapped with (case I vs case II) and were excluded for better readability. Torque errors were not normally distributed in all cases and increased during WF, WD, and fast EF motions. 

Inter-case (II-IV) error differences within repetitions were small but significant ($p < 0.05$). Intra-repetition differences, comparing errors of each case with itself, were negligible. Intra-object error differences (across all repetitions) for each case with itself were significant, $p < 0.05$, except for WD in the Can/Drill scenario. 

Figs. \ref{fig: torque error} (b) and (c) compare Case I (grey-scale, dashed) and Case II (coloured) torque trajectories during the first repetition. Additional dotted lines mark prosthesis-related torque terms for the same cases. Case I $\boldsymbol{\tau}_{P}$ was obtained in physical experiments without objects, and Case II $\boldsymbol{\tau}_{P_s}$ in simulation by \eqref{equ: equations of motion} without object wrench. Computed and measured torque trajectories correlated well, except for wrist WF and WD joints during their respective motions. There $\boldsymbol{\tau}_{P_s}$ $\cong$ 0 Nm diverged from measured $\boldsymbol{\tau}_{P}$, with the latter exhibiting similar trajectories as in trials including objects. Fast elbow motion produced torque spikes in the EF and WD joints.\begingroup
\setlength{\tabcolsep}{5pt} 
\begin{table}[ht]
\caption{Joint Torque statistics - measured against computed, and percentile torque contributions of objects averaged over three repetitions per trial. Torques are in Nm.}
\vspace{-1\baselineskip}
\label{tab: Stat_TorqueTraj}
\begin{tabular}{ll|lllllll}
Object & Joint &           Avg &          $\overline{Max}$ &          $\overline{RMS}$ & $\overline{Corr}$ &         $\overline{p95}$ &       $\overline{p50}$ &         $\overline{p5}$ \\ \hline
\multirow{4}{*}{Can} &    EF &  0.27 & 0.92 & 0.28 & 0.98 &   35 &  34 &  34 \\
 &    PS & -0.11 & 0.26 & 0.17 & 0.96 &   73 &  80 &  58 \\
 &    WF & -0.08 & 0.39 & 0.15 & 0.98 &   46 &  78 &  44 \\
 &    WD & -0.01 & 0.35 & 0.18 & 0.96 &   62 &  45 &  45 \\ \hline
\multirow{4}{*}{Bottle} &    EF &  0.28 & 0.90 & 0.29 & 0.98 &   37 &  36 &  36 \\
 &    PS & -0.13 & 0.24 & 0.17 & 0.93 &   71 &  82 & -18 \\
 &    WF & -0.09 & 0.39 & 0.16 & 0.98 &   48 &  79 &  47 \\
 &    WD & -0.01 & 0.42 & 0.19 & 0.96 &   15 &  47 &  48 \\ \hline
\multirow{4}{*}{Drill} &    EF &  0.22 & 0.99 & 0.24 & 0.97 &   44 &  44 &  43 \\
 &    PS & -0.15 & 0.18 & 0.19 & 0.96 &   70 &  83 &  43 \\
 &    WF & -0.11 & 0.51 & 0.22 & 0.98 &   61 &  78 &  59 \\
 &    WD & -0.01 & 0.74 & 0.30 & 0.93 & -121 &  57 &  59 \\
\end{tabular}
\end{table}
\endgroup

Table \ref{tab: Stat_TorqueTraj} reports statistical values of Case I against Case II errors, averaged over three repetitions per trial. Standard deviations are omitted due to their small magnitudes. In comparison to peak torques plotted in Figs \ref{fig: torque error} (b) and (c), average and \ac{RMS} torque errors are low, while maximal errors reach 10-20\% of peak magnitudes. Pearson correlation is high at 0.93 to 0.98 for all joints and objects ($p << 0$). Percentile object torques were between 34\% and 80\% of limb torques, where influences were generally higher for 50th percentile torques. Percentages were computed as limb percentile over object percentile torques.

\section{Discussion} 
\label{sec: Discusion}
The \ac{OAA} was developed for enhancing motion-captured manipulation trials of human subjects with virtual objects $V_{obj}$. Use cases include augmenting existing \ac{adl} databases with derived object trajectories and tracking items unfit for sensor or optical marker placement. Resulting object motions can, for example, inform about object-induced joint torques during the manipulation of objects, which is rarely reported in the literature. Recording dedicated experiments that are time and resource-consuming is not required. Instead, tapping into vast data resources allows for automatically investigating varying object dynamics. \\

The chosen input marker set consists of two anatomic wrist markers $M_w$ and five markers placed on the dorsal side of the hand $M_h$ (minimum three markers). The wrist styloids can be accurately located by their bony landmarks with associated markers commonly recorded in motion capture experiments \cite{ganguly2021comparison, ganguly2021can, metcalf2008validation}. Consequently, the vector pointing from the \ac{US} to the \ac{RS} marker is a reliable source for the $\vec{p}$ hand frame component. In return, the algorithm requisites the wrist deviation angle, which requires inverse kinematics of subject joints before object augmentation. However, the alternative of a similar marker pair for the $\vec{p}$ direction located on the hand, which intrinsically contains the wrist angle, is suboptimal. This is because of the general unavailability of such markers in accessible datasets. Moreover, the computation of the same direction from other hand markers cannot be reliably automated due to the imprecise nature of marker placement. When less than three hand markers are available, $\vec{n}$ can be computed from the union of hand and wrist markers. However, resulting $\vec{n}$ are rotated around $\vec{p}$, compared to $\vec{n}$ produced by the \ac{OAA}.\\

The performance of the \ac{OAA} was experimentally validated on the 4\ac{dof} \ac{ANP}  platform \cite{AlexanderToedtheide2022_MIRMI_transhumeralProsthesis}, equipped with accurate torque sensors in each joint and using a gold standard optical motion capture system. 
Three unique objects were grasped with a 5 \ac{dof} robotic anthropomorphic hand. The system was moved along 4 pre-programmed joint motions. Motions were designed to test each joint in isolation. The experimental protocol involved three repetitions per object, with objects being re-grasped between each repetition. Grasping manoeuvres were executed without strict adherence to maintaining consistent relative configurations between the object and the hand. Object selection was deliberate, aiming to expose the limitations of the algorithm: the diameter of the Can was at the limits of the robotic hand-grasping capabilities; the elongated shape of the bottle accentuated tracking errors of $VM_t$; and the forward-facing geometry of the drill led to potential misalignment issues around the $\vec{p}$ axis.\\

The $M_h$ cluster in this study consisted of five optical markers. However, the number and placement of $M_h$ markers vary between databases and subjects, influencing the location and orientation of computed hand frames. An offset was, therefore, added to the computed VM to model the effects of variations in $M_h$. The offset direction was chosen to $\vec{f}$ since it maximised the changes in object-wrench moment arms for all joints except PS. 
The offset represents the extreme scenario where markers are placed randomly, which is rarely the case, especially for biomechanical grasping experiments. We did not incorporate variations in the orientation of hand coordinate frames due to the curvature of the dorsal side of the human hand, as well as skin stretching, which could lead to fluctuations in the estimated direction of $\vec{n}$ based on a limited number of markers, into our analysis.\\

Virtual markers were positioned within hand coordinate frames through transformations relying exclusively on object dimensions. This approach mirrors real-world scenarios where object trajectories are unavailable and grasping procedures lack sufficient documentation for precise replication. OpenSim guidelines recommend maximal and root mean square (RMS) marker tracking errors to be below 20 mm and 10 mm, respectively, for static poses. Our findings consistently fell well below these thresholds during static and dynamic motions. One exception occurred with the Drill's virtual marker trajectories ($VM_t$) during supination, attributed to the neglected rotation around the $\vec{p}$ axis for the Drill. Virtual object trajectories exhibited a high correlation with motion-captured ground truth trajectories. Wrist deviation phases were performed with the forearm pronated at 90$^{\circ}$, alleviating torque limitations of the \ac{ANP}s wrist by reducing gravitational joint loads. Forearm pronation generally increased marker errors and was the reason for increased tracking errors during wrist deviation phases. This led to relatively large error deviations for both can and drill $VM_t$s, reflecting grip variations expected in trials with human subjects. \\

Joint torques were computed with object wrenches from virtual, virtual + offset, and ground truth object trajectories. During these scenarios, the variations of object placement within hand frames had negligible influence on torque trajectories and distributions. As a result, joint torques produced by the \ac{OAA} are robust to variations in hand marker locations while justifying the placement of objects by their dimensions. Maximal torque errors reached 20\% of peak measured joint torques and occurred in the wrist WF and WD joints during their respective motion phases. However, in the absence of objects, wrist torques were expected to be nearly 0 since motions were slow and gravity was aligned with the joint axes. In contrast, Figs. \ref{fig: torque error} (b) and (c) show nearly overlapping torque spikes for trials with and without experimental objects. Consequently, deviations between measured and computed wrist torque trajectories were caused by unmodelled effects within the tendon-driven wrist mechanism. Similar effects were observed in the elbow joint during fast elbow motion. Forearm pronation and supination motions caused vibrations in the robotic hand, translating to added noise in joint torque measurements. As a result, torque errors exhibit added noise during forearm rotation motion intervals. Special care has to be taken when the grasp between the object and hand is not firm. Object dynamics causing location and orientation changes in the hand frame were not considered. This effect was noticed during the fast elbow motion phase of the Drill, where $VM_t$ errors exhibit local maxima. \\

The existence and nature of physical objects utilised in the experiments must be considered during object augmentation. When $V_{obj}$ inertia and mass properties are different to the experimental conditions, object influences are not accurately reflected by the available limb motion of human subjects. Our experiments did not show such effects since joint motions were pre-programmed and tracked by the prosthesis controller set to high stiffness. However, the presence of objects during manipulation experiments cannot be neglected, as demonstrated by the substantial contribution of object wrenches to joint torques (table \ref{tab: Stat_TorqueTraj}). Additionally, \cite{Akhavanfar2022_2HandSimCompare} demonstrated that interaction wrench-based approaches, assuming accurate object motions, reduce dynamic inconsistencies compared to simply adding object dynamics or gravity-oriented loads to the hand. The latter two options are inapplicable in scenarios where objects are supported and/or constrained by the environment (e.g., doors, sliding over surfaces, etc.). In contrast, knowledge about object motions allows the computation of interaction wrenches, considering possible environmental influences, while adding object mass and inertia to the hand still requires estimating the object-to-hand configuration. \\

We successfully employed the \ac{OAA} to augment motions pertaining to object manipulation of the ADL Human Arm Motion Dataset \cite{Gloumakov2021_ADL_dataset}. Joint quantities were computed for a wide range of objects, enabled by the \ac{OAA}'s ability to compute prescribed motions and induced joint torques for arbitrary objects. In addition, the \ac{OAA} allowed the simulation of joint torques produced by small or kinematically constrained items, such as turning a key in a lock and opening a door. Respective results are presented in \cite{hernethFunctionalKinematicKinetic2024}. Other databases, such as \cite{Liang2020_ASIAN_ADL, SenayMihcin2023_ADL_PrevioslyExcludedADL} may be augmented similarly, enriching the information on \ac{adl} joint kinetics available in literature.\\

In conclusion, the algorithm enables the automatic computation of hand-coordinate frames from optical hand and wrist markers. It is robust to variations in the number and placement of hand markers, ensuring its repeatability between trials and subjects. Objects can be placed within the hand-coordinate systems by homogeneous transformations based on object dimensions. Object motions and object-induced joint torques demonstrated satisfactory tracking of measured ground truth trajectories. It provides a dependable tool for enhancing existing databases with virtual objects while enabling accurate tracking of objects unfit for marker or sensor placement.

\section{Appendix} 
\label{sec: Appendix}
An open source implementation of the \ac{OAA} is available at \url{https://github.com/ChristopherHerneth/ObjectAugmentationAlgorithm/tree/main}. Accompanying video material may be found here \url{https://youtu.be/8oz-awvyNRA}.

\begin{acronym}
\acro{ANP}[ANP]{Artificial Neuromuscular Prosthesis}
\acro{LMC}[LMC]{Leap Motion Controller}
\acro{ADLDAT}[ADL Dataset]{ADL Human Arm Motion Dataset}
\acro{IQR}[IQR]{Interqartile range}
\acro{FRoM}[FRoM]{Functional Range of Motion}
\acro{RoM}[RoM]{Range of Motion}
\acro{CoM}[CoM]{Center of Mass}
\acro{dof}[DoF]{Degree of Freedom}
\acro{adl}[ADL]{activities of daily living}
\acro{dulm}[MoBL-ARMS DULm]{MoBL-ARMS Dynamic Upper Limb model}
\acro{tpm}[TPm]{Transhumeral Prosthesis model}
\acro{SVD}[SVD]{Singular Value Decomposition}
\acro{RMS}[RMS]{Root Mean Square}
\acro{MoCap}[MoCap]{Motion capture}
\acro{OAA}[OAA]{Object Augmentation Algorithm}
\acro{IK}[IK]{Inverse Kinematics}
\acro{ID}[ID]{Inverse Dynamics}
\acro{RS}[RS]{Radial Styloid}
\acro{US}[US]{Ulnar Styloid}
\acro{VM}[VM]{Virtual Marker}
\end{acronym}


\section*{ACKNOWLEDGMENT}
This work was supported by the Federal Ministry of Education and Research of the Federal Republic of Germany (BMBF) by funding the project AI.D under Project Number 16ME0539K.

\typeout{}
\bibliography{Bibliography}

\begin{thebibliography}{10}

\bibitem{li2023gendexgrasp}
Puhao Li, Tengyu Liu, Yuyang Li, Yiran Geng, Yixin Zhu, Yaodong Yang, and Siyuan Huang.
\newblock Gendexgrasp: Generalizable dexterous grasping.
\newblock In {\em 2023 IEEE International Conference on Robotics and Automation (ICRA)}. IEEE, May 2023.

\bibitem{wang2023dexgraspnet}
Ruicheng Wang, Jialiang Zhang, Jiayi Chen, Yinzhen Xu, Puhao Li, Tengyu Liu, and He~Wang.
\newblock Dexgraspnet: A large-scale robotic dexterous grasp dataset for general objects based on simulation.
\newblock In {\em 2023 IEEE International Conference on Robotics and Automation (ICRA)}. IEEE, May 2023.

\bibitem{Olugbade2022_HumanMovementDatabases}
Temitayo Olugbade, Marta Bieńkiewicz, Giulia Barbareschi, Vincenzo D’amato, Luca Oneto, Antonio Camurri, Catherine Holloway, Mårten Björkman, Peter Keller, Martin Clayton, Amanda C De~C Williams, Nicolas Gold, Cristina Becchio, Benoît Bardy, and Nadia Bianchi-Berthouze.
\newblock Human movement datasets: An interdisciplinary scoping review.
\newblock {\em ACM Computing Surveys}, 55(6):1--29, December 2022.

\bibitem{akhbari2019predicting}
Bardiya Akhbari, Douglas~C. Moore, David~H. Laidlaw, Arnold‐Peter~C. Weiss, Edward Akelman, Scott~W. Wolfe, and Joseph~J. Crisco.
\newblock Predicting carpal bone kinematics using an expanded digital database of wrist carpal bone anatomy and kinematics.
\newblock {\em Journal of Orthopaedic Research}, 37(12):2661--2670, August 2019.

\bibitem{jarque2020large}
{Scientific Data Curation Team}.
\newblock Metadata record for: A large calibrated database of hand movements and grasps kinematics, 2020.

\bibitem{Rosena_ADL_torque}
J.~Rosen, J.C. Perry, N.~Manning, S.~Burns, and B.~Hannaford.
\newblock The human arm kinematics and dynamics during daily activities - toward a 7 dof upper limb powered exoskeleton.
\newblock In {\em ICAR ’05. Proceedings., 12th International Conference on Advanced Robotics, 2005.} IEEE, 2005.

\bibitem{Toedtheide2023_ExoPro}
Alexander Toedtheide, Edmundo~Pozo Fortunić, Johannes Kühn, Elisabeth~Rose Jensen, and Sami Haddadin.
\newblock A wearable force-sensitive and body-aware exoprosthesis for a transhumeral prosthesis socket.
\newblock {\em IEEE Transactions on Robotics}, 39(3):2203--2223, June 2023.

\bibitem{Buongiorno2018}
Domenico Buongiorno, Edoardo Sotgiu, Daniele Leonardis, Simone Marcheschi, Massimiliano Solazzi, and Antonio Frisoli.
\newblock Wres: A novel 3 dof wrist exoskeleton with tendon-driven differential transmission for neuro-rehabilitation and teleoperation.
\newblock {\em IEEE Robotics and Automation Letters}, 3(3):2152--2159, July 2018.

\bibitem{Fennel2022}
Michael Fennel, Antonio Zea, and Uwe~D. Hanebeck.
\newblock Optimization-driven design of a kinesthetic haptic interface with human-like capabilities.
\newblock {\em IEEE Transactions on Haptics}, 15(1):45--50, January 2022.

\bibitem{Sha2020}
Liansen Sha, Andi Lin, Qiang Xi, and Shaolong Kuang.
\newblock A topology optimization method for robot light-weight design under multi-working conditions and its application on upper-limb powered exoskeleton.
\newblock In {\em 2020 International Conference on Artificial Intelligence and Electromechanical Automation (AIEA)}. IEEE, June 2020.

\bibitem{Adams2015}
Richard~J. Adams, Matthew~D. Lichter, Eileen~T. Krepkovich, Allison Ellington, Marga White, and Paul~T. Diamond.
\newblock Assessing upper extremity motor function in practice of virtual activities of daily living.
\newblock {\em IEEE Transactions on Neural Systems and Rehabilitation Engineering}, 23(2):287--296, March 2015.

\bibitem{Udawatta2019}
Lanka Udawatta, Mohammad Alwedian, Simon Hissem, Bryan Daniels, and Mouza Almazrouei.
\newblock Analyzing angular velocity of the human arm in rhythmic walking.
\newblock In {\em 2019 Advances in Science and Engineering Technology International Conferences (ASET)}. IEEE, March 2019.

\bibitem{Damerla2022_2Rod_Parallel}
Revanth Damerla, Kevin Rice, Daniel Rubio-Ejchel, Maurice Miro, Enrico Braucher, Juliet Foote, Issam Bourai, Aaryan Singhal, Kang Yang, Hongju Guang, Vasil Iakimovitch, Evelyn Sorgenfrei, and Shorya Awtar.
\newblock Design and testing of a novel, high-performance two {DoF} prosthetic wrist.
\newblock {\em {IEEE} Transactions on Medical Robotics and Bionics}, 4(2):502--519, may 2022.

\bibitem{AlexanderToedtheide2022_MIRMI_transhumeralProsthesis}
Alexander Toedtheide, Edmundo~Pozo Fortonic, Johannes Kuehn, Elisabeth Jensen, and Sami Haddadin.
\newblock A transhumeral prosthesis with an artificial neuromuscular system: Sim2real-guided design, modeling and control.
\newblock {\em International Journal of Robotics Research}, 2022.

\bibitem{Sasaki2010_WT(angle)}
Makoto Sasaki, Takehiro Iwami, Kazuto Miyawaki, Ikuro Sato, Goro Obinata, and Ashish Dutt.
\newblock Higher dimensional spatial expression of upper limb manipulation ability based on human joint torque characteristics.
\newblock In {\em Robot Manipulators New Achievements}. {InTech}, apr 2010.

\bibitem{Biddiss2007_25Years_abbandonment_studyreview}
Elaine~A. Biddiss and Tom~T. Chau.
\newblock Upper limb prosthesis use and abandonment.
\newblock {\em Prosthetics and Orthotics International}, 31(3):236--257, sep 2007.

\bibitem{Davidson2002_Austr_Prost_Use}
Judith Davidson.
\newblock A survey of the satisfaction of upper limb amputees with their prostheses, their lifestyles, and their abilities.
\newblock {\em Journal of Hand Therapy}, 15(1):62--70, jan 2002.

\bibitem{Cordella2016_Need_ULPU}
Francesca Cordella, Anna~Lisa Ciancio, Rinaldo Sacchetti, Angelo Davalli, Andrea~Giovanni Cutti, Eugenio Guglielmelli, and Loredana Zollo.
\newblock Literature review on needs of upper limb prosthesis users.
\newblock {\em Frontiers in Neuroscience}, 10, may 2016.

\bibitem{Akhavanfar2022_2HandSimCompare}
Mohammadhossein Akhavanfar, Thomas~K. Uchida, Allison~L. Clouthier, and Ryan~B. Graham.
\newblock Sharing the load: modeling loads in {OpenSim} to simulate two-handed lifting.
\newblock {\em Multibody System Dynamics}, 54(2):213--234, jan 2022.

\bibitem{Muller2022_pointMass_torqueestimation}
Antoine Muller, Hakim Mecheri, Philippe Corbeil, André Plamondon, and Xavier Robert-Lachaine.
\newblock Inertial motion capture-based estimation of l5/s1 moments during manual materials handling.
\newblock {\em Sensors}, 22(17):6454, August 2022.

\bibitem{Krebs2021_ADL_KIT_Bimanual}
Franziska Krebs, Andre Meixner, Isabel Patzer, and Tamim Asfour.
\newblock The kit bimanual manipulation dataset.
\newblock {\em IEEE/RAS International Conference on Humanoid Robots (Humanoids)}, 2021.

\bibitem{Mandery2015_ADL_KIT_WholeBody}
Christian Mandery, Oemer Terlemez, Martin Do, Nikolaus Vahrenkamp, and Tamim Asfour.
\newblock The kit whole-body human motion database.
\newblock {\em International Conference on Advanced Robotics (ICAR)}, 2015.

\bibitem{Liang2020_ASIAN_ADL}
Phyllis Liang, Wai~Hang Kwong, Ananda Sidarta, Choon~Kong Yap, Wee~Kiat Tan, Lek~Syn Lim, Pui~Yee Chan, Christopher Wee~Keong Kuah, Seng~Kwee Wee, Karen Chua, Colin Quek, and Wei~Tech Ang.
\newblock An asian-centric human movement database capturing activities of daily living.
\newblock {\em Scientific Data}, 7(1), September 2020.

\bibitem{SenayMihcin2023_ADL_PrevioslyExcludedADL}
{Şenay Mihçin}, Ahmet~Mert Sahin, Mehmet Yılmaz, Alican~Tuncay Alpkaya, Merve Tuna, Nuray~Korkmaz Can, Serap Sahin, Sevinc Akdeniz, and Aliye Tosun.
\newblock Database covering the previously excluded daily life activities, 2023.

\bibitem{OuadoudiBelabzioui2023_upperLimbJointLoadANN}
Hasnaa Ouadoudi~Belabzioui, Charles Pontonnier, Georges Dumont, Pierre Plantard, and Franck Multon.
\newblock {\em Estimation of Upper-Limb Joint Torques in Static and Dynamic Phases for Lifting Tasks}, pages 71--80.
\newblock Springer Nature Switzerland, 2023.

\bibitem{Wang2021L5/S1MomentEstimationDuringLifting}
Hanwen Wang, Ziyang Xie, Lu~Lu, Li~Li, Xu~Xu, and Edward~P. Fitts.
\newblock A single-camera computer vision-based method for 3d l5/s1 momentestimation during lifting tasks.
\newblock {\em Proceedings of the Human Factors and Ergonomics Society Annual Meeting}, 65(1):472--476, September 2021.

\bibitem{Delp2007_OpenSim}
Scott~L. Delp, Frank~C. Anderson, Allison~S. Arnold, Peter Loan, Ayman Habib, Chand~T. John, Eran Guendelman, and Darryl~G. Thelen.
\newblock {OpenSim}: Open-source software to create and analyze dynamic simulations of movement.
\newblock {\em {IEEE} Transactions on Biomedical Engineering}, 54(11):1940--1950, nov 2007.

\bibitem{AboulHagag2011_handsize}
Khaled~E. Aboul-Hagag, Soheir~A. Mohamed, Maha~A. Hilal, and Eman~A. Mohamed.
\newblock Determination of sex from hand dimensions and index/ring finger length ratio in upper egyptians.
\newblock {\em Egyptian Journal of Forensic Sciences}, 1(2):80--86, June 2011.

\bibitem{ganguly2021comparison}
Amartya Ganguly, Gabriel Rashidi, and Katja Mombaur.
\newblock Comparison of the performance of the leap motion controllertm with a standard marker-based motion capture system.
\newblock {\em Sensors}, 21(5):1750, March 2021.

\bibitem{ganguly2021can}
Amartya Ganguly, Gabriel Rashidi, and Katja Mombaur.
\newblock Can leap motion controller replace conventional marker-based motion capture systems?
\newblock {\em XXVIII Congress of the International Society of Biomechanics}, 2021.

\bibitem{metcalf2008validation}
C.D. Metcalf, S.V. Notley, P.H. Chappell, J.H. Burridge, and V.T. Yule.
\newblock Validation and application of a computational model for wrist and hand movements using surface markers.
\newblock {\em IEEE Transactions on Biomedical Engineering}, 55(3):1199--1210, March 2008.

\bibitem{Gloumakov2021_ADL_dataset}
Yuri Gloumakov, Adam Spiers, and Aaron Dollar.
\newblock Adl human arm motion data.
\newblock IEEE Dataport: 10.21227/ayhf-q263, 2021.

\bibitem{hernethFunctionalKinematicKinetic2024}
Christopher Herneth, Amartya Ganguly, and Sami Haddadin.
\newblock Functional kinematic and kinetic requirements of the upper limb during activities of daily living: A recommendation on necessary joint capabilities for prosthetic arms, August 2024.

\end{thebibliography}

\addtolength{\textheight}{-12cm}   

\end{document}